\documentclass[sigconf]{acmart}

\usepackage{booktabs} 
\usepackage{todonotes}
\usepackage{tabu}
\usepackage[ruled]{algorithm2e}
\usepackage{soul}
\usepackage{algorithmic}
\usepackage{color, colortbl}
\definecolor{Gray}{gray}{0.9}
\definecolor{light-gray}{gray}{0.95}

\usepackage{hyperref}

\copyrightyear{2019}
\acmYear{2019}
\setcopyright{acmcopyright}
\acmConference[GECCO '19]{Genetic and Evolutionary Computation Conference}{July 13--17, 2019}{Prague, Czech Republic}
\acmBooktitle{Genetic and Evolutionary Computation Conference (GECCO '19), July 13--17, 2019, Prague, Czech Republic}
\acmPrice{15.00}
\acmDOI{10.1145/3321707.3321875}
\acmISBN{978-1-4503-6111-8/19/07}

\begin{document}
\title{Lexicase Selection of Specialists}


\author{Thomas Helmuth}
\orcid{0000-0002-2330-6809}
\affiliation{%
  \institution{Hamilton College}
  \city{Clinton} 
  \state{New York} 
  \country{USA}
}
\email{thelmuth@hamilton.edu}

\author{Edward Pantridge}
\orcid{0000-0003-0535-5268}
\affiliation{%
  \institution{Swoop, Inc.}
  \city{Cambridge} 
  \state{Massachusetts} 
  \country{USA}
}
\email{ed@swoop.com}

\author{Lee Spector}
\orcid{0000-0001-5299-4797}
\affiliation{%
	\institution{Hampshire College}
	\city{Amherst}
	\state{Massachusetts}
	\country{USA}
}
\email{lspector@hampshire.edu}

\begin{abstract}
Lexicase parent selection filters the population by considering one random training case at a time, eliminating any individuals with errors for the current case that are worse than the best error in the selection pool, until a single individual remains. This process often stops before considering all training cases, meaning that it will ignore the error values on any cases that were not yet considered. Lexicase selection can therefore select specialist individuals that have poor errors on some training cases, if they have great errors on others and those errors come near the start of the random list of cases used for the parent selection event in question. We hypothesize here that selecting these specialists, which may have poor total error, plays an important role in lexicase selection's observed performance advantages over error-aggregating parent selection methods such as tournament selection, which select specialists much less frequently. We conduct experiments examining this hypothesis, and find that lexicase selection's performance and diversity maintenance degrade when we deprive it of the ability of selecting specialists. These findings help explain the improved performance of lexicase selection compared to tournament selection, and suggest that specialists help drive evolution under lexicase selection toward global solutions.
\end{abstract}

%
%



\maketitle

\section{Introduction} \label{section:intro}

Most parent selection methods used in genetic programming, and in genetic algorithms more generally, select individuals on the basis of scalar fitness values. For problems that involve multiple training cases, these fitness values are aggregated over all of the training cases, often by summing them. By contrast, lexicase selection selects parents on the basis of performance on un-aggregated training-case errors \cite{Spector:2012:GECCOcompA,Helmuth:2015:ieeeTEC,LaCava:EC}. It does this by considering training cases one at a time, in a different random order for each parent selection event. For each parent selection event it creates a pool that initially contains the entire population, and then for each training case, it filters the pool to retain only the individuals with the best error for each training case. If the pool is  reduced to a single individual, then that individual is the selected parent. If many individuals survive filtering by all of the training cases, then a randomly chosen survivor is designated as the selected parent.

Prior work has shown that lexicase selection often works well in practice, but the reasons that it does so, and the contexts in which it does and doesn't work well, are still topics of active investigation. In the present paper we address one hypothesis regarding the efficacy of lexicase selection: that selecting \textit{specialists} is important for solving problems.
By ``specialists'' we mean individuals with relatively low errors on a subset of of the training cases but high errors on other training cases and subsequently poor total error relative to the rest of the population. In contrast to specialists, generalists perform approximately the same on all training cases, not doing particularly well on any training cases while having overall good total error.

Our motivation for the present study stems from anecdotal evidence observed in an earlier study, which suggested that specialists might contribute in important ways to the evolution of solutions \cite{McPhee:2015:GPTP}. This prior work also suggested that the selection of specialists might explain, to a significant degree, the better problem-solving performance of lexicase selection relative to other parent selection methods. 

More specifically, in this prior work we examined the lineage leading to a solution to the ``Replace Space with Newline'' software synthesis problem, evolved with a PushGP genetic programming system. In the run that we examined, the generation in which a solution first appeared actually contained 45 distinct solutions. All of these solutions were children of the same parent in the previous generation, and both this parent and and {\it its} parent (that is, the grandparent of all of the solutions) had total error values that were in the worst quartile of their respective generations by total error. The grandparent of every solution had nearly the worst total error of its generation. Nonetheless, both the grandparent and the parent produced large numbers of offspring, including large numbers of solutions in the final generation.

A later study using a larger set of benchmark problems observed lexicase selection selecting individuals with high total error significantly more frequently than tournament selection~\cite{Pantridge:2018:GECCO:SEL}. This study also observed that lexicase selection rarely utilizes a majority of the training cases when selecting parents.

These observations motivated the present study, but anecdotal evidence is not sufficient to ground scientific understanding or to guide engineering practice. Systematic studies are required to determine the extent to which the selection of specialists is truly important, and the contexts in which this is the case. In this paper we document such a study, providing the first clear evidence supporting the hypothesis that the selection of specialists is responsible, in large measure, for the superiority of lexicase selection to tournament selection.

In the following sections we present background on lexicase selection and then the design, results, and analysis of our new experiments.

\section{Background on Lexicase Selection}
\label{section:lexicase-background}


The basic and most commonly used version of the lexicase selection algorithm proceeds as follows each time a parent is required:

\begin{enumerate}
    \item A collection of {\bf candidates} is set initially to contain the entire population.
    \item A collection of {\bf cases} is set initially to contain all of the training cases, shuffled in random order.
    \item Until a parent has been designated, loop:
    \begin{enumerate}
    \item Discard all individuals in {\bf candidates} except those with exactly the lowest error for the first case in {\bf cases}.
    \item If just a single individual remains in {\bf candidates}, then designate it as the parent.
    \item If only a single item remains in {\it cases}, then designate a randomly chosen individual from {\bf candidates} as the parent.
    \item Otherwise, remove the first item from {\bf cases}.
    \end{enumerate}
\end{enumerate}

Lexicase selection has been studied in several settings, and several variants of the basic algorithm have been proposed (for example, \cite{10.1007/978-3-319-90512-9_7}). Among the most significant of these variations is epsilon lexicase selection, in which ``exactly the lowest error'' in the description of the algorithm is replaced with ``within epsilon of the lowest error'' for a suitably defined epsilon; this has proven to be particularly effective on problems with floating-point errors \cite{LaCava:2016:GECCO,LaCava:EC}. Additionally, lexicase selection has been effectively used to solve problems in areas such as boolean logic and finite algebras~\cite{Helmuth:2015:ieeeTEC, Helmuth:2013:GECCOcomp, Liskowski:2015:GECCOcomp}, evolutionary robotics~\cite{moore:2017:ecal}, and boolean constraint satisfaction using genetic algorithms~\cite{Metevier2019}.

Lexicase selection often produces and maintains particularly diverse populations, and this has been hypothesized to be responsible, in part, for its problem-solving power \cite{Helmuth:2015:GPTP, Helmuth:2016:GECCOcomp}. If lexicase selection does in fact select specialists more often than other parent selection techniques, this may contribute to its effects on diversity, regardless of effects on problem-solving performance.

Populations evolving by lexicase selection are also often observed to exhibit {\it hyperselection}, in which single individuals in one generation are used as parents for many, sometimes most or nearly all, of the children in the next generation. The causal connections between hyperselection and problem-solving power are complex \cite{Helmuth:2016:GECCO}, but in any case this may also be relevant to the interpretation of experimental results on specialist selection, since the presence or absence of specialists may influence the frequency and patterns of hyperselection.

An additional aspect of lexicase selection that bears consideration is the fact that selected individuals will always be nondominated in their populations and elite with respect to at least one training case, a property that has been characterized as inhabiting the ``corners'' of the Pareto front \cite{LaCava:EC}. This too should be considered in the interpretation of results on specialist selection.

\section{Specialists in Genetic Programming}

A specialist is an individual that achieves low errors on a subset of training cases while having high errors on other training cases. The total, or aggregated, error of a specialist individual is often relatively high compared to the rest of the population, since a poor error on a few training cases can dominate the sum of the errors. In contrast to specialists, a generalist is an individual that performs approximately the same on all training cases, achieving neither particularly good nor particularly poor results on any training case, and often achieving relatively good total error. Consider the following training cases for the function $y = (x_1)^2 - x_2$.

\begin{figure}[H]
    \centering
    \begin{tabular}{cc|c}
        $x_1$ & $x_2$ & $y$  \\ \hline
        2 & 1 & 3 \\
        3 & 5 & 4 \\
        1 & 3 & -2
    \end{tabular}
\end{figure}

The following two tables describe the actual output ($\hat{y}$) and expected output ($y$) of a generalist and a specialist on each training case.

\begin{figure}[H]
    \centering
    \begin{tabular}{cc|c}
        \multicolumn{3}{c}{Generalist} \\
        $\hat{y}$ & $y$ & Error  \\ \hline
        10 & 3 & 7 \\
        -8 & 4 & 12 \\
        6.5 & -2 & 8.5 \\ \hline
        \multicolumn{2}{r}{Total:} & 27.5
    \end{tabular}
    \quad
    \begin{tabular}{cc|c}
        \multicolumn{3}{c}{Specialist} \\
        $\hat{y}$ & $y$ & Error  \\ \hline
        -100 & 3 & 103 \\
        \texttt{err} & 4 & 1,000,000 \\ 
        -1.99 & -2 & 0.01  \\ \hline
        \multicolumn{2}{r}{Total:} & 1,000,103.01
    \end{tabular}
\end{figure}

The generalist has similar error values across all training cases while the specialist has a near zero error on one training case but high errors on the other training cases.\footnote{On an actual problem with many training cases, a specialist will likely perform well on a subset of the training cases, not just one of them.}
Notice that the specialist has received a penalty error of one million on the second training case because it could not be evaluated on the given set of inputs. 

The total error of the specialist is drastically higher than the generalist. However, the generalist was not able to achieve a near zero error on any of the training cases. In an evolutionary population that is ranked by total error, the generalists will tend to have lower rank than the specialists. On the other hand, the specialist may have discovered something truly useful about solving the problem as indicated by its one (or more with real problems) nearly perfect output, and might be worth selecting to pass on its genetics to the next generation.

\section{Experimental design}
\label{section:experimental-design}

In Section~\ref{section:intro} we described a single run that featured an individual in the bottom quartile of the population (when sorted by total error) that was the parent of 45 solution programs. Later, in Section~\ref{section:lexicase-specialists} we will show that specialists make up large portions of the individuals selected by lexicase selection compared to tournament selection. Still, this does not answer the question of whether selecting specialists is an important component to lexicase selection's improved performance compared to tournament selection and other selection methods, or whether it is a side effect that has little bearing on the trajectory of evolution.

Does lexicase selection perform well because it selects specialists, or can it maintain good performance without selecting individuals with poor total error? We hypothesize that lexicase selection's ability to select specialist individuals with poor total error allows it to more effectively explore the search space than if it were limited to selecting individuals with good performance when measured by total error.
We do not expect tournament selection to exhibit similar decreases in performance when limited to selecting individuals with good total error, since it does not often select individuals with poor total error.
Additionally, we expect that limiting lexicase selection to individuals with better total error will decrease population diversity.

To test our hypotheses, we propose an experiment where parent selection cannot select individuals with poor total error relative to the population. We devised a new survival selection step to run before parent selection called \textit{elitist survival selection}. During elitist survival selection, we sort the population by total error and only allow the best $X\%$ of the population to ``survive'' to be available to make children. We call the percent of the population that survives this step the \textit{elitist survival rate}. We then conduct parent selection using this reduced population as normal. With 100\% elitist survival we would keep the entire population (i.e. no individuals are removed); 30\% elitist survival would keep only the best 300 individuals sorted by total error (out of a population of 1000) to be available for parent selection. If our hypothesis holds, we would expect to see decreased performance with lexicase selection but not with tournament selection.

\subsection{Benchmark Problems}

The problems used in the experiments described here were taken from a benchmark suite of software synthesis problems, which were derived from exercises in introductory computer science textbooks \cite{Helmuth:2015:GECCO}. These problems require general-purpose programming to solve, such as multiple data types (strings, integers, floats, Booleans, vectors, etc.) and various control flow techniques.
These problems have been addressed in several studies, using multiple genetic programming systems using lexicase selection including PushGP~\cite{Helmuth:2015:GECCO, Helmuth:thesis, McPhee:2015:GPTP, Helmuth:2015:GPTP, Helmuth:2016:GECCO, Helmuth:2016:GECCOcomp, Helmuth:2017:GECCO, McPhee:2017:GECCOa} and grammar guided GP~\cite{Forstenlechner:2018:CEC, Forstenlechner:2017:EuroGP, Forstenlechner:2018:PPSN, Forstenlechner:2018:GECCO}, as well as at least one non-evolutionary program synthesis technique~\cite{DBLP:journals/corr/abs-1811-10665}.

We selected 8 out of the 29 benchmark problems to use in this study to reflect a wide range of requirements and difficulties.
The specific problems addressed in this study are Last Index of Zero, Mirror Image, Negative to Zero, Replace Space with Newline, String Lengths Backwards, Syllables, Vector Average, and X-Word Lines.
Some of these problems have been solved with genetic programming using lexicase selection over 75 times out of 100, while others have solution rates around 25\%.

In this study, we follow the lead of the benchmark suite in how to determine whether a run is successful or not~\cite{Helmuth:2015:GECCO}. Each GP run uses a different randomly-generated set of training cases, as well as a larger set of unseen test cases used to assess generalization. Once a program has evolved that passes all of the training cases, we test it on the unseen test set---if it passes those as well, it counts as a solution. In this paper we additionally automatically simplify the programs that pass the training data before testing them for generalization, a process that shrinks program size without changing the behavior of the program on the training set. Previous work has shown that automatic simplification effectively increases generalization on these benchmark problems~\cite{Helmuth:2017:GECCO}.

\subsection{Push and PushGP}

The experiments conducted in this study were run using a PushGP genetic programming system, which evolves stack-based programs expressed in the Push programming language \cite{spector:2002:GPEM, 1068292, Pantridge:2017:GECCO}. 
The key feature of Push for the experiments presented here is its multi-stack architecture, which includes a stack for each data type and instructions that always take their arguments from the correct stacks and push their results to the correct stacks. This facilitates the evolution of programs that use multiple, nontrivial data and control structures, making it suitable for solving the benchmark problems described above. In addition, a wealth of prior data on the performance of PushGP on these problems can provide context for the results obtained in different experimental conditions \cite{Helmuth:thesis,McPhee:2015:GPTP,Helmuth:2015:GPTP,Helmuth:2016:GECCO,Helmuth:2016:GECCOcomp,Helmuth:2017:GECCO,McPhee:2017:GECCOa}. We use the Clojure implementation of PushGP\footnote{\url{https://github.com/lspector/Clojush}}, which was also used in the aforementioned studies.

\begin{table}
\centering
\caption{PushGP system parameters and the usage rates of genetic operators.}
\label{table:PushGP-params}
\begin{tabular}{l r}
\toprule
\textbf{Parameter} & \textbf{Value} \tabularnewline
\midrule
population size & 1000 \tabularnewline
max number of generations & 300 \tabularnewline
tournament size for tournament selection & 7 \tabularnewline
\midrule
\textbf{Genetic Operator Rates} & \textbf{Prob} \tabularnewline
\midrule
alternation & 0.2 \tabularnewline
uniform mutation & 0.2 \tabularnewline
uniform close mutation & 0.1 \tabularnewline
alternation followed by uniform mutation & 0.5 \tabularnewline
\bottomrule
\end{tabular}
\end{table}

The parameters and configurations of the PushGP system that we used for the experiments here are the same as those described in the original benchmark description~\cite{Helmuth:2015:GECCO}. Table~\ref{table:PushGP-params} presents the key parameters. The version of the code used in our experiments is made available here: \url{https://github.com/thelmuth/Clojush/releases/tag/GECCO-Lexicase-Selection-Of-Specialists}.

\section{Specialists Under Tournament Selection}
\label{section:specialists-under-tourney}

Tournament selection displays an inherent pressure to select generalists due to its utilization of an aggregate error metric, such as RMSE, classification accuracy, or total error. To compute these kinds of error metrics, an individual's errors on all training cases must be considered. If an individual performs particularly poorly on any subset of training cases, its aggregated error will be raised and probability of getting selected will decrease.

\begin{figure}[t] 
\centering
\includegraphics[width=\linewidth]{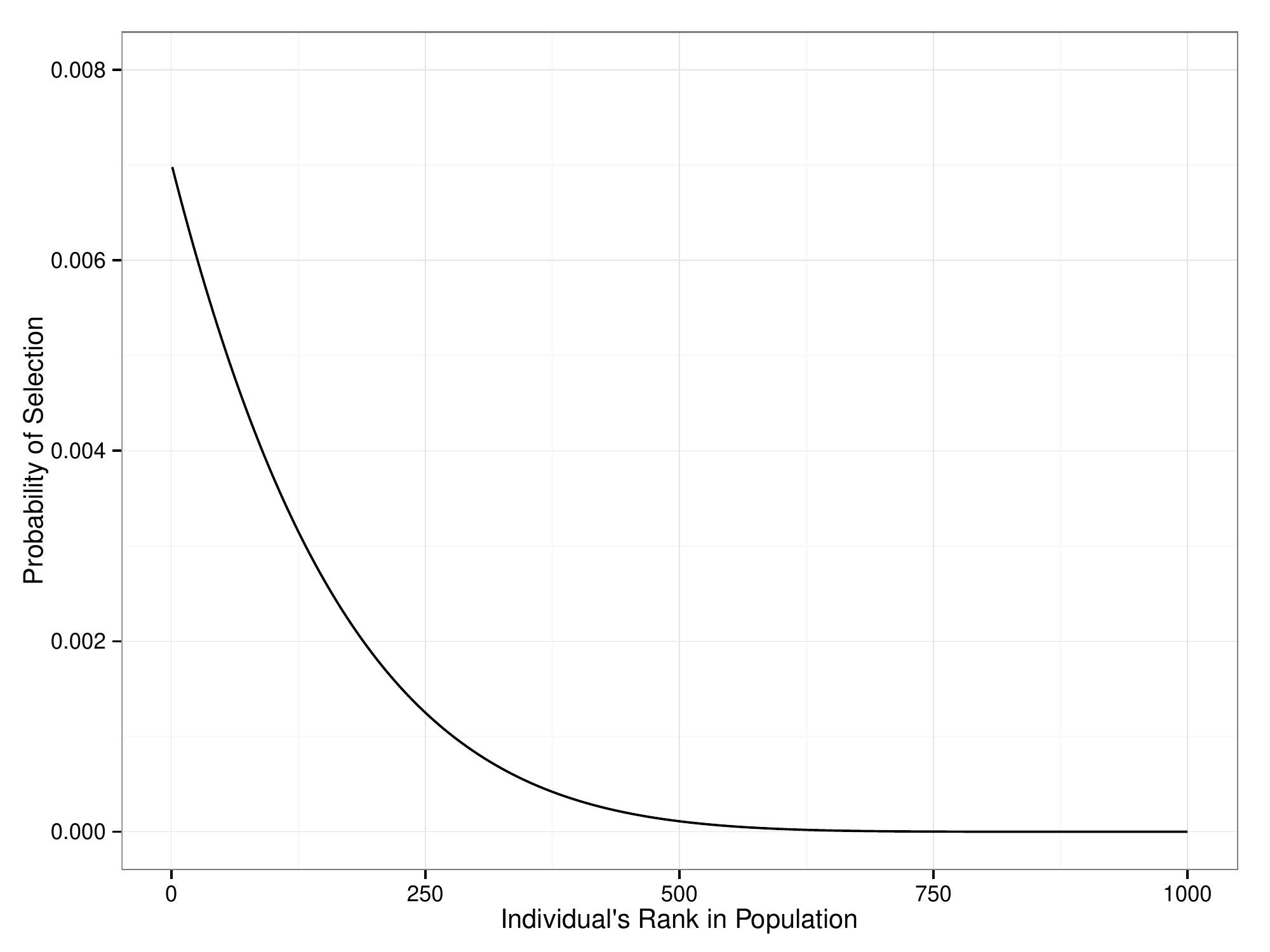}
\caption{Probability mass function of selecting individual with rank $i$ out of a population of 1000 individuals using tournament selection with tournament size 7, assuming no two individuals have the same rank. This plots Equation~\ref{tourneyProbEquation}.}
\label{fig:prob-selection-tourney-7}
\end{figure}

\begin{table}[t]
\centering
\caption{Theoretical probability of tournament selection selecting an individual that would be removed by $X\%$ elitist survival. For example, the probability of selecting an individual removed by 50\% elitist survival is 0.00781, meaning that individuals with total error worse than the median make up less than 0.8\% of the parents when using tournament selection.}
\label{table:prob-select-tourney-elitist-survival}
\begin{tabu} to 2.3in {X[1,r] X[2.7,r]}
\toprule
\textbf{\% Elitist Survival} & \textbf{Probability of Selecting A Removed Individual} \\
\midrule
10                  & 0.47829    \\
20                  & 0.20971    \\
30                  & 0.08235    \\
40                  & 0.02799    \\
50                  & 0.00781    \\
60                  & 0.00163    \\
70                  & 0.00021    \\
80                  & 0.00001    \\
90                  & 0.0000001    \\
100                 & 0    \\
\bottomrule                       
\end{tabu}
\end{table}

With tournament selection, the number of times an individual can be selected is limited by the number of tournaments in which it participates. If the best member of the population participates in 1\% of the tournaments for a given generation, it will be selected up to 1\% of the time that generation, but no more.
 Since the expected number of tournaments in which each individual participates is constant for a particular population size $P$ and tournament size $t$, the probability of an individual being selected by tournament selection is entirely determined by its rank in the population. In particular, B\"ack \cite{350042, Blickle:1995:MAT:645514.658088} shows that the probability of selecting an individual with rank $i \in [1,P]$, with $i = 1$ being the best rank, is
\begin{equation}\label{tourneyProbEquation}
p(i) = \frac{(P-i+1)^t - (P-i)^t}{P^t}
\end{equation}
assuming no two individuals have the same fitness. With ties in the rankings, this equation does not hold exactly, but is approximately correct unless there are many tied individuals. We plot this probability mass function in Figure~\ref{fig:prob-selection-tourney-7}.

In experiments without elitist survival, tournament selection selected individuals in the worst 50\% of the population (by total error) at a rate of 3.3\%. This is greater than the 0.78\% predicted by the theoretical probability of selection due to the selected benchmark problems producing a high numbers of ties in total error.

Table \ref{table:prob-select-tourney-elitist-survival} shows that tournament selection rarely selects poor-ranking individuals. Both theoretical and empirical evidence suggest that tournament selection will almost never select specialists. Thus, the elitist survival filtering should not have a strong impact on tournament selection's ability to find solution programs.

\section{Specialists Under Lexicase Selection} \label{section:lexicase-specialists}

\begin{figure*} 
    \centering
    \includegraphics[width=0.8\linewidth]{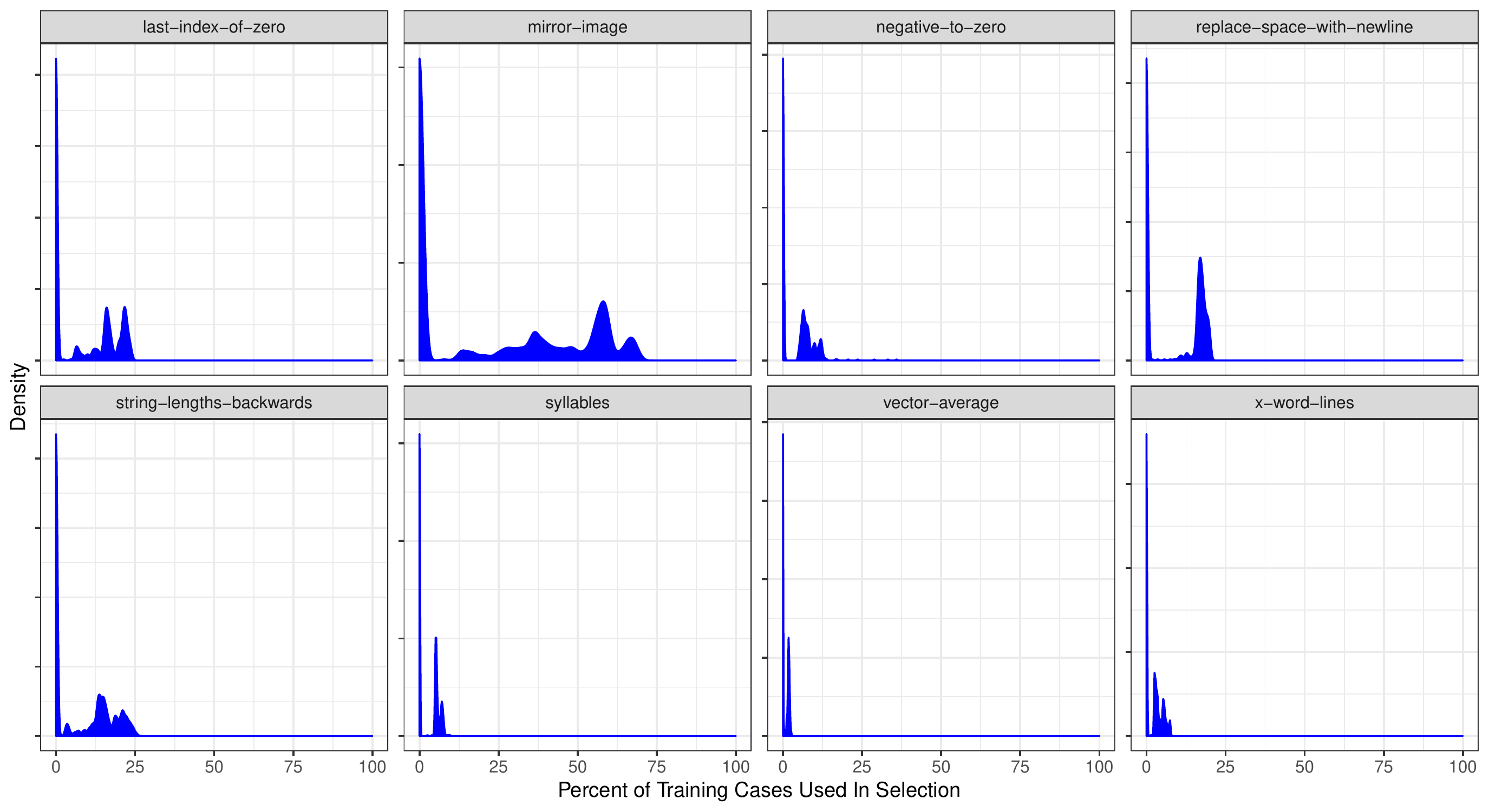}
    \caption{Each plot shows the distribution of the percent of the total training cases used in each lexicase selection event. With the exception of the Mirror Image problem, lexicase selection almost never considers more than one third of the training cases. This implies that selected individuals can have arbitrarily high errors on the training cases not considered during any given selection event. }
\label{figure:es:case-usage-distrib}
\end{figure*}

By considering training cases one at a time, lexicase selection often selects an individual without considering all of the training cases; this idea explicitly influenced the design of lexicase selection. When halting before seeing all of the training cases, the lexicase algorithm will ignore the error values on all other training cases, regardless of whether they are relatively good or relatively poor compared to the rest of the population. Lexicase selection therefore has the ability to select specialist individuals that perform extremely well on some cases while having very poor error on other cases.

Figure \ref{figure:es:case-usage-distrib} plots the distribution of the number of training cases used in each selection event across each problem.
It should be noted that this statistic clearly varies between problems and that these results assume a ``preselection'' phase that reduces collections of individuals that have identical errors on all training cases to one randomly selected member; otherwise, any time multiple individuals have the same error vector, 100\% of the training cases would be used, since none of them would differentiate the individuals.

Figure \ref{figure:es:case-usage-distrib} shows that in practice, lexicase selection rarely considers more than 50\% of the training cases, and often less than 25\%. These measurements agree with the empirical results on a different set of benchmarks obtained in a previous study~\cite{Pantridge:2018:GECCO:SEL}. These results provide evidence that many of lexicase's selections ignore 50-75\% of the training cases; it is certainly possible that some of these selected individuals achieved poor errors on some of the ignored training cases.

Based on the experimental results of this paper, lexicase selection selects individuals in the worst 50\% of the population (by total error) at a rate of 7.9\%. Although this may seem low, it is more than twice the rate of tournament selection, as discussed in section \ref{section:specialists-under-tourney}.

\begin{figure*} 
    \centering
    \includegraphics[width=.85\textwidth]{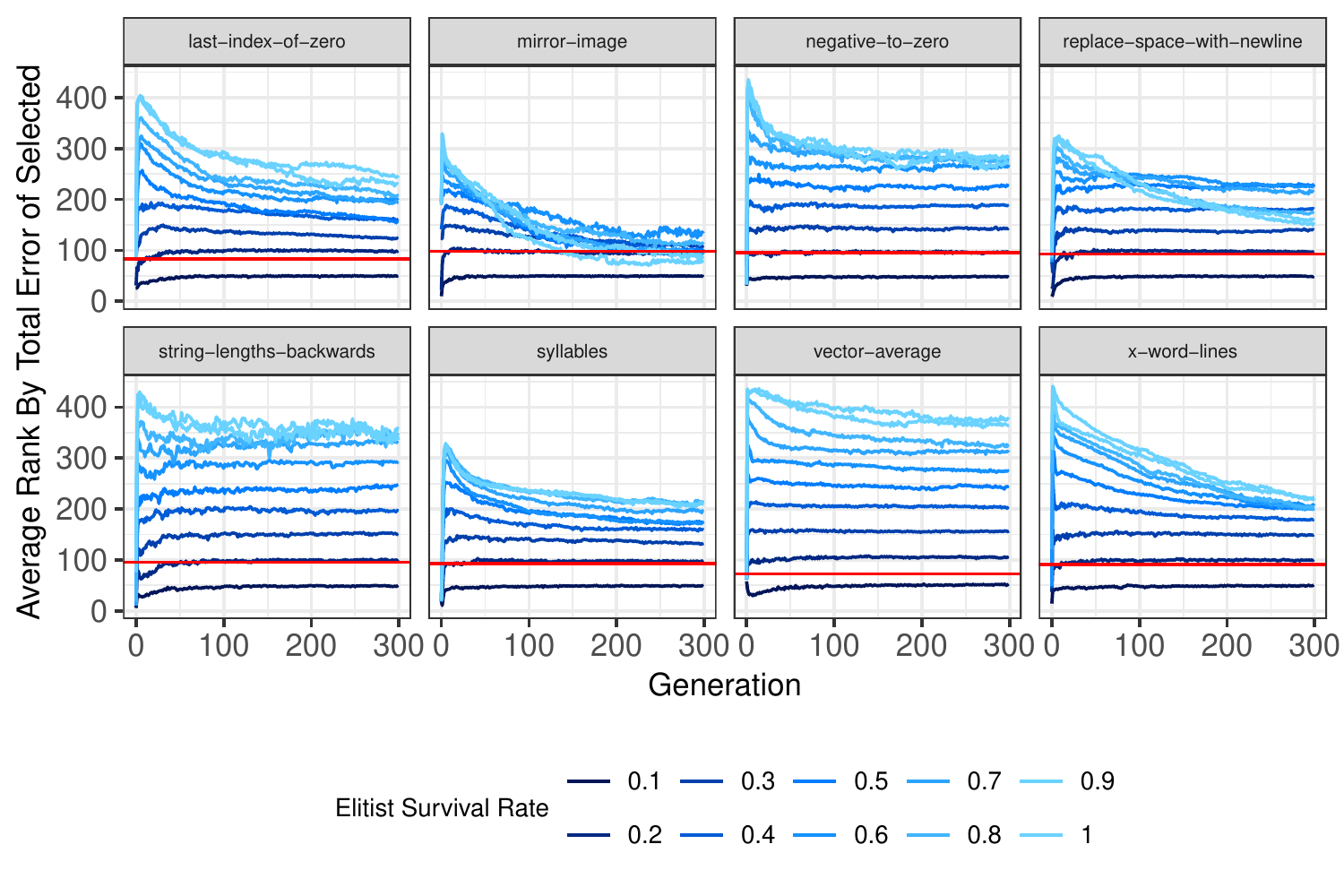}
    \caption{The average rank of individuals selected at each generation by lexicase selection using various elitist survival filters. The red horizontal lines show the overall median rank of individuals selected by tournament selection with no elitist survival filtering.}
    \label{figure:es:avg-rank-by-gen}
\end{figure*}

\begin{figure*} 
\centering
\includegraphics[width=.8\linewidth]{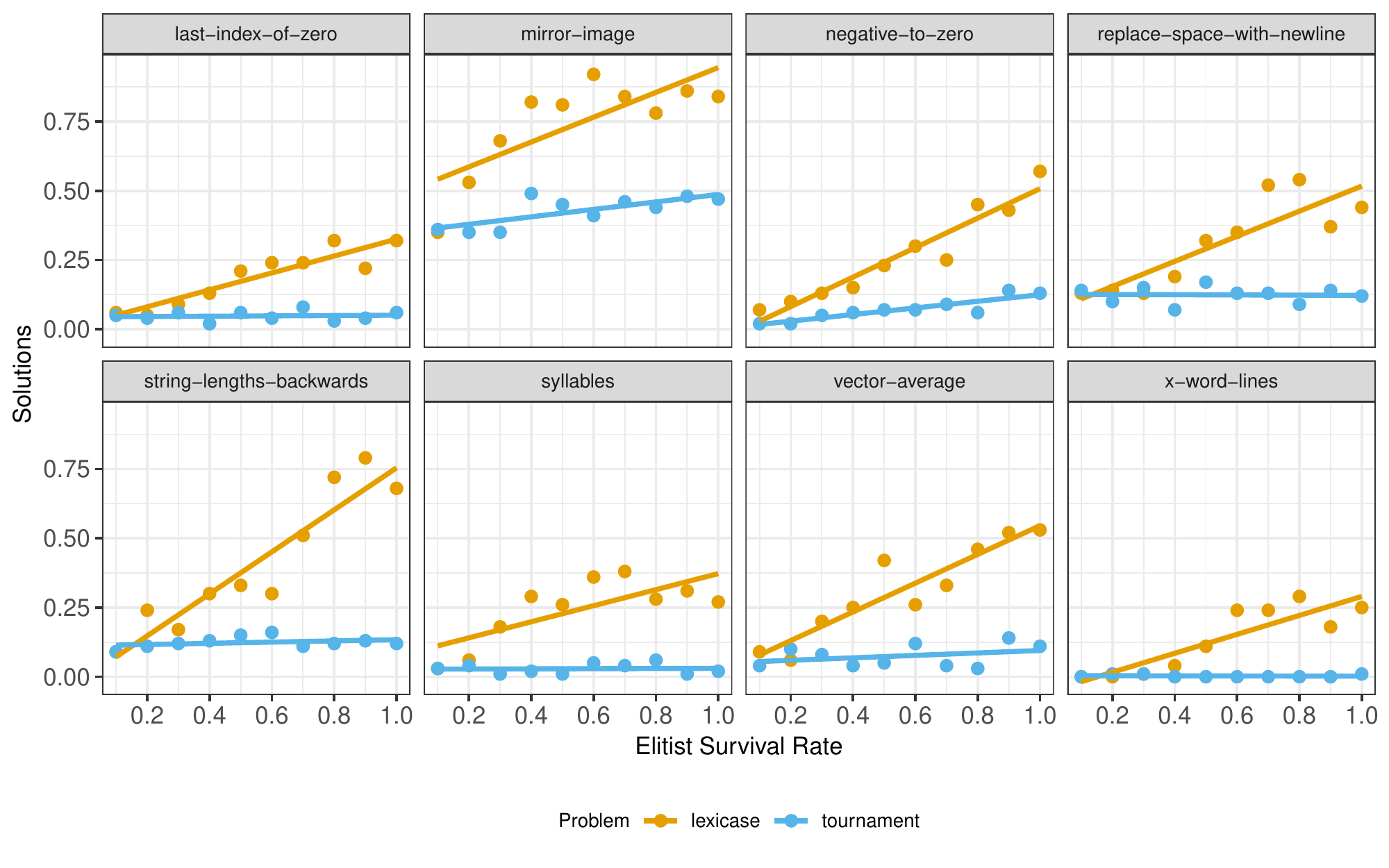}
\caption{The impact of elitist survival filtering on the ability of lexicase selection and tournament selection to find generalizing solution programs. As the elitist filtering is loosened, lexicase selection tends to find more solutions. For every problem, an $F$-test was used to determine that the relationship between the elitist survival rate and the number of solutions when using lexicase selection is significant at the 0.05 level. This same relationship for tournament selection was shown to be not significant at the 0.05 level for all problems.}
\label{figure:es:solutions-by-rate}
\end{figure*}

Figure \ref{figure:es:avg-rank-by-gen} shows how the average rank of individuals selected by lexicase selection changes throughout evolution. Lexicase selection with no elitist survival filtering (or where elitist survival rate is 100\%) begins evolution by selecting individuals with very high rank, especially compared to tournament selection. As evolution searches the space, lexicase selection tends to select individuals with lower ranks. The average rank of individuals selected by lexicase selection is rarely lower than the average rank of individuals selected by tournament selection. 

For many of the problems in Figure~\ref{figure:es:avg-rank-by-gen}, the average rank of selected individuals slowly decreases throughout evolution, especially for higher levels of elitist survival rate. This indicates that lexicase turns more toward low-rank (low total error) individuals later in runs. While we are not sure exactly what causes this phenomenon, we hypothesize that in many cases, lexicase selection has concentrated on one part of the search space, attempting to refine one or more promising programs into solutions (and likely often doing so).

Since lexicase selection only retains individuals with elite errors on the first test case it considers, every selected individual is elite on a subset of the training cases. In combination with the observed tendency to select individuals with high total error ranks, it can be concluded that lexicase selection is definitively selecting specialists at a much higher rate than tournament selection.

\section{Importance of Selecting Specialists}
\label{section:poor-performing-specialists}


In Section~\ref{section:experimental-design}, we made the hypothesis that lexicase selection's ability to select specialist individuals with poor total error improves its performance compared to if it were limited to selecting individuals with good total error. To test this hypothesis, we conduct runs of PushGP
using elitist survival with elitist survival rates of 10\% to 100\% in increments of 10. By using elitist survival, we can force selection (lexicase or tournament) to not select individuals with total error worse than some percent of the population. Thus, for example, if it is important for lexicase to select specialist individuals with rank (when sorted by total error) worse than 60\% of the population, then we would expect lexicase selection with 60\% elitist survival to perform worse than with 100\% survival of the population.

Based on Equation~\ref{tourneyProbEquation}, we can calculate how often we would expect tournament selection to select the individuals excluded by elitist survival.\footnote{Figure~\ref{fig:prob-selection-tourney-7}, which plots the probability distribution defined by Equation~\ref{tourneyProbEquation}, is useful when visualizing these cumulative probabilities.}
The probabilities of tournament selection choosing an individual removed by elitist survival at different rates are given in Table~\ref{table:prob-select-tourney-elitist-survival}. 
Tournament selection would select a decent proportion of the individuals removed by 30\% elitist survival, at around 0.08. We can see that most of those individuals have ranks between 30\% and 50\%, since tournament selection selects individuals worse than the median with probability of only about 0.0078. Thus, we would not expect 50\% survival elitism to affect the performance of tournament selection, and certainly not 70\% survival elitism. Even 20\% survival elitism may have negligible effects.

Figure~\ref{figure:es:solutions-by-rate} gives the number of successful runs on 8 benchmark problems using elitist survival rates of 10\% to 90\% in increments of 10 with lexicase and tournament selection. We compare these results to 100\% elitist survival, which is equivalent to not using elitist survival, since the entire population is kept. We plot a linear regression line for each problem, and use an $F$-test to determine if there is a relationship between the elitist survival rate and the number of solutions for each method. 

On all 8 of the problems, there is a significant relationship between the elitist survival rate and the number of solutions found by lexicase selection, indicating that lexicase performs significantly worse when limited to smaller elite proportions of the population. On the other hand, none of the problems showed a significant relationship between elitist survival rate and number of solutions when using tournament selection. As predicted by the small effects of lower-rank individuals on tournament selection (as discussed in Section~\ref{section:specialists-under-tourney}), removing those lower-rank individuals has little effect on the performance of tournament selection. In fact, limiting tournament selection to only the top 10\% of the population by total error gave numbers of successes insignificantly different from using the entire population on every problem except for Negative to Zero (using a chi-squared test).

\begin{figure}[t] 
\centering
\includegraphics[width=\linewidth]{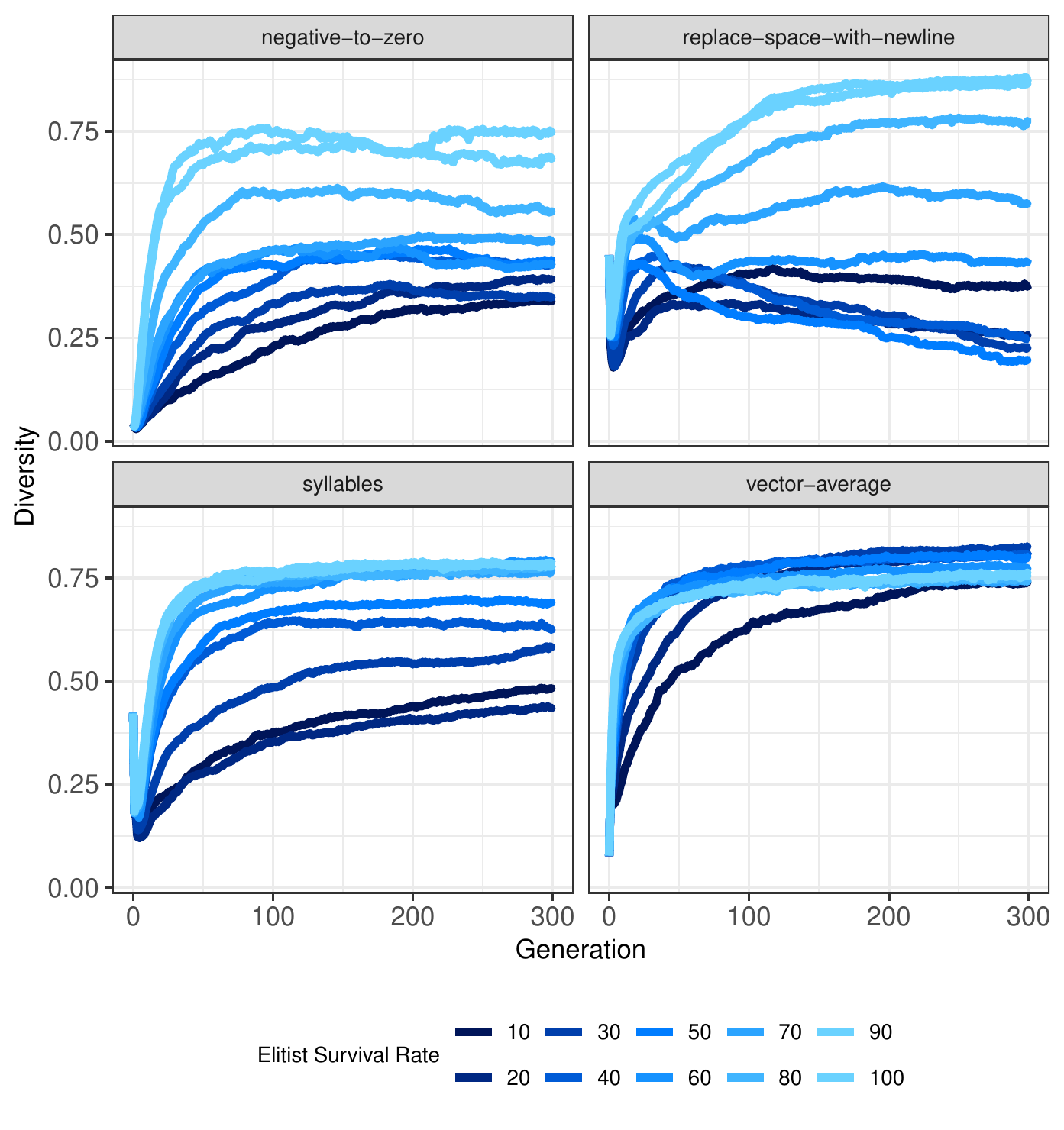}
\caption{Mean behavioral diversity for lexicase selection using elitist survival at different rates on four representative problems.}
\label{figure:es:diversity-overview}
\end{figure}

The \textit{behavioral diversity} of a population is the proportion of distinct behavior vectors that are present in a population, where a \textit{behavior vector} is simply the outputs of a program when run on the training cases~\cite{Jackson:2010:PPSN}.
We plot the population behavioral diversity of lexicase selection for four of the problems in Figure~\ref{figure:es:diversity-overview}.
We do not present the diversities of populations under tournament selection, which were very low for every problem and every elitist survival rate, mirroring to previous studies comparing the behavioral diversity of lexicase and tournament selection~\cite{Helmuth:2015:GPTP, Helmuth:2016:GECCOcomp}. This result is consistent with the unchanged performance of tournament selection with elitist survival, both of which can be explained by the small portion of selections affected by elitist survival.

For the first three problems in Figure~\ref{figure:es:diversity-overview}, the behavioral diversity of runs using lexicase selection decreases as the number of individuals removed by elitist survival increases. We see this decrease across all problems besides Vector Average, though the impact varies per problem. On the Replace Space With Newline problem, lexicase selection with elitist survival rates below 80\% grow in diversity early, but then lose diversity and finish the remainder of the run with low levels of diversity. On the other Negative to Zero and Syllables problems, the lower rates of elitist survival have similar curves to the higher rates, just at lower levels.

One interesting finding here is that for the Vector Average problem, removing the worst individuals \textit{increases} behavioral diversity, down to 30\% elitist survival. This strange pattern says that if lexicase selection is given fewer individuals to select, the resulting populations will be more diverse. Despite the higher levels of diversity, Vector Average followed a similar pattern of poorer performance with low levels of elitist survival, as shown in Figure~\ref{figure:es:solutions-by-rate}.

These results provide evidence supporting our hypothesis that lexicase selection makes use of specialist individuals with poor total error relative to the rest of the population---individuals that presumably have poor errors on some training cases but good errors on others. Lexicase selection shows clear correlation between elitist survival and success rate on every problem, performing better when able to select from the worst individuals when sorted by total error. Since the individuals lexicase selects from these parts of the population have poor total error yet receive selections, they must perform well, or even be elite, on some training cases. Lexicase selection performs better when allowed to select these specialists, which clearly help drive the direction of evolution toward more solutions.

Our plots show that behavioral diversity in lexicase selection runs decreases, sometimes significantly, as we decrease the number of individuals that survive elitist survival. These plots support our hypothesis that the high diversity seen in runs using lexicase selection is influenced by lexicase selection's ability to select individuals with relatively poor total error. Selecting specialists thus allows lexicase selection to better explore the search space, likely contributing to its better performance. The decreased diversity in runs using lower rates of elitist survival likely contributes to the corresponding decreases in performance observed in Figure~\ref{figure:es:solutions-by-rate}.

As expected, tournament selection was not significantly affected by elitist survival selection at any level, even when removing 90\% of the population. As we saw in Table~\ref{table:prob-select-tourney-elitist-survival}, tournament selection simply does not often select specialist parents from the bottom ranks of the population. Instead, it concentrates on the individuals with the best total error, and mostly selects from the top 20\% of the population, with more than half of selections coming from individuals in the top 10\% of the population. This difference between lexicase selection and tournament selection explains at least part of their difference in performance as seen in previous studies~\cite{Helmuth:2015:GECCO, Helmuth:2015:ieeeTEC}.


\section{Conclusions}

Numerous demonstrations of lexicase selection's search performance have concluded it is superior to tournament selection on a variety of tasks. Previous attempts to explain this behavior have noted observations of increased population diversity, guarantees of non-dominated selections, and the possibility of selecting individuals with high total error. This paper formalizes the hypothesis that lexicase selection's performance is in part due to its tendency to select specialists over generalists, especially compared to tournament selection. These specialists, with excellent errors on some training cases yet poor total error, receive little attention from most other parent selection methods, which aggregate performance into a single fitness metric.

This paper presents theory explaining the exceedingly low probability of tournament selection selecting specialists, along with empirical results that support this theory. In contrast, we observe the comparatively high rate that lexicase selection selects specialists. We additionally provide evidence of test case usage during lexicase selection, indicating that few test cases are typically used in any one parent selection event, providing evidence for how lexicase can select specialists by ignoring training cases on which the specialist performed poorly.

To support the hypothesis that the selection of specialists is a key component of lexicase selection's search performance, an elitist survival filter was applied with various degrees of strictness before conducting parent selection. This filtering removed all potential specialists and forced lexicase selection to select among more generalist individuals, which have better total error. The filter significantly reduced the number of solutions evolution was able to find, implying that the presence of specialists was crucial to lexicase selection's performance. Furthermore, tournament selection was not significantly impacted by elitist survival filtering. Additionally, we discussed the effects of specialists on a population's diversity under lexicase selection, finding that specialists typically contribute to lexicase selection's ability to maintain high rates of population diversity.


The work presented here helps focus the understanding of what particular mechanics of lexicase selection drive its noted performance improvements over other parent selection techniques. Earlier work that showed that while lexicase selection hyperselects individuals at high rates, this hyperselection does not significantly impact the performance of GP using lexicase selection~\cite{Helmuth:2016:GECCO}. Here, we illuminate one aspect of lexicase selection that appears to be key to its success, namely the selection of specialists typically ignored by other parent selection techniques.

These findings suggest that future work to improve parent selection techniques should consider their ability to select specialist individuals, which provided significant benefits to lexicase selection in this study. Additionally, we solely focused on the automatic program synthesis domain here; programs in this domain can use control flow to act in different modalities for different inputs~\cite{Spector:2012:GECCOcompA}, potentially leading to the development and importance of specialists. It could prove informative to replicate this study in other domains, especially ones with relatively small instruction sets such as symbolic regression.

How does selecting specialists lead to solving problems? How are the skills of specialists adapted or combined into better individuals? These questions go beyond the selection of parents and depend on how the GP system generates children from specialist parents. A solution program must by definition be a generalist, since it perfectly passes all of the test cases. Future work should consider how generalists are constructed from specialists, and if there are better ways of doing so.

The specialists selected by lexicase selection here were subjected to the same genetic operators used in other studies with PushGP. However, we could imagine designing genetic operators with specialists in mind to better make use of their novel abilities. For example, when combining two specialists, should we use a different recombination operator than when combining two generalists, to have a better chance at reaping the benefits of both parents? We could imagine such operators increasing the efficiency of evolution as it combines the specializations of individuals until it finds a general solution.

\begin{acks}
We thank Nicholas McPhee and members of the Hampshire College Institute for Computational Intelligence for discussions that advanced this work. This material is based upon work supported by the National Science Foundation under Grant No. 1617087. Any opinions, findings, and conclusions or recommendations expressed in this publication are those of the authors and do not necessarily reflect the views of the National Science Foundation.
\end{acks}

\bibliographystyle{ACM-Reference-Format}
\bibliography{lex-specialists} 


\begin{thebibliography}{29}


\ifx \showCODEN    \undefined \def \showCODEN     #1{\unskip}     \fi
\ifx \showDOI      \undefined \def \showDOI       #1{#1}\fi
\ifx \showISBNx    \undefined \def \showISBNx     #1{\unskip}     \fi
\ifx \showISBNxiii \undefined \def \showISBNxiii  #1{\unskip}     \fi
\ifx \showISSN     \undefined \def \showISSN      #1{\unskip}     \fi
\ifx \showLCCN     \undefined \def \showLCCN      #1{\unskip}     \fi
\ifx \shownote     \undefined \def \shownote      #1{#1}          \fi
\ifx \showarticletitle \undefined \def \showarticletitle #1{#1}   \fi
\ifx \showURL      \undefined \def \showURL       {\relax}        \fi
\providecommand\bibfield[2]{#2}
\providecommand\bibinfo[2]{#2}
\providecommand\natexlab[1]{#1}
\providecommand\showeprint[2][]{arXiv:#2}

\bibitem[\protect\citeauthoryear{B{\"a}ck}{B{\"a}ck}{1994}]%
        {350042}
\bibfield{author}{\bibinfo{person}{Thomas B{\"a}ck}.}
  \bibinfo{year}{1994}\natexlab{}.
\newblock \showarticletitle{Selective pressure in evolutionary algorithms: a
  characterization of selection mechanisms}. In \bibinfo{booktitle}{{\em
  Evolutionary Computation, 1994. IEEE World Congress on Computational
  Intelligence., Proceedings of the First IEEE Conference on}}.
  \bibinfo{pages}{57--62 vol.1}.
\newblock
\showDOI{%
\url{https://doi.org/10.1109/ICEC.1994.350042}}


\bibitem[\protect\citeauthoryear{Blickle and Thiele}{Blickle and
  Thiele}{1995}]%
        {Blickle:1995:MAT:645514.658088}
\bibfield{author}{\bibinfo{person}{Tobias Blickle} {and}
  \bibinfo{person}{Lothar Thiele}.} \bibinfo{year}{1995}\natexlab{}.
\newblock \showarticletitle{A Mathematical Analysis of Tournament Selection}.
  In \bibinfo{booktitle}{{\em Proceedings of the 6th International Conference
  on Genetic Algorithms}}. \bibinfo{publisher}{Morgan Kaufmann Publishers
  Inc.}, \bibinfo{address}{San Francisco, CA, USA}, \bibinfo{pages}{9--16}.
\newblock
\showISBNx{1-55860-370-0}
\showURL{%
\url{http://dl.acm.org/citation.cfm?id=645514.658088}}


\bibitem[\protect\citeauthoryear{Forstenlechner, Fagan, Nicolau, and
  O'Neill}{Forstenlechner et~al\mbox{.}}{2017}]%
        {Forstenlechner:2017:EuroGP}
\bibfield{author}{\bibinfo{person}{Stefan Forstenlechner},
  \bibinfo{person}{David Fagan}, \bibinfo{person}{Miguel Nicolau}, {and}
  \bibinfo{person}{Michael O'Neill}.} \bibinfo{year}{2017}\natexlab{}.
\newblock \showarticletitle{A Grammar Design Pattern for Arbitrary Program
  Synthesis Problems in Genetic Programming}. In \bibinfo{booktitle}{{\em
  EuroGP 2017: Proceedings of the 20th European Conference on Genetic
  Programming}} {\em (\bibinfo{series}{LNCS})},
  \bibfield{editor}{\bibinfo{person}{Mauro Castelli}, \bibinfo{person}{James
  McDermott}, {and} \bibinfo{person}{Lukas Sekanina}} (Eds.),
  Vol.~\bibinfo{volume}{10196}. \bibinfo{publisher}{Springer Verlag},
  \bibinfo{address}{Amsterdam}, \bibinfo{pages}{262--277}.
\newblock
\showDOI{%
\url{https://doi.org/10.1007/978-3-319-55696-3_17}}


\bibitem[\protect\citeauthoryear{Forstenlechner, Fagan, Nicolau, and
  O'Neill}{Forstenlechner et~al\mbox{.}}{2018a}]%
        {Forstenlechner:2018:PPSN}
\bibfield{author}{\bibinfo{person}{Stefan Forstenlechner},
  \bibinfo{person}{David Fagan}, \bibinfo{person}{Miguel Nicolau}, {and}
  \bibinfo{person}{Michael O'Neill}.} \bibinfo{year}{2018}\natexlab{a}.
\newblock \showarticletitle{Extending Program Synthesis Grammars for
  Grammar-Guided Genetic Programming}. In \bibinfo{booktitle}{{\em 15th
  International Conference on Parallel Problem Solving from Nature}} {\em
  (\bibinfo{series}{LNCS})}, \bibfield{editor}{\bibinfo{person}{Anne Auger},
  \bibinfo{person}{Carlos~M. Fonseca}, \bibinfo{person}{Nuno Lourenco},
  \bibinfo{person}{Penousal Machado}, \bibinfo{person}{Luis Paquete}, {and}
  \bibinfo{person}{Darrell Whitley}} (Eds.), Vol.~\bibinfo{volume}{11101}.
  \bibinfo{publisher}{Springer}, \bibinfo{address}{Coimbra, Portugal},
  \bibinfo{pages}{197--208}.
\newblock
\showDOI{%
\url{https://doi.org/10.1007/978-3-319-99253-2_16}}


\bibitem[\protect\citeauthoryear{Forstenlechner, Fagan, Nicolau, and
  O'Neill}{Forstenlechner et~al\mbox{.}}{2018b}]%
        {Forstenlechner:2018:GECCO}
\bibfield{author}{\bibinfo{person}{Stefan Forstenlechner},
  \bibinfo{person}{David Fagan}, \bibinfo{person}{Miguel Nicolau}, {and}
  \bibinfo{person}{Michael O'Neill}.} \bibinfo{year}{2018}\natexlab{b}.
\newblock \showarticletitle{Towards effective semantic operators for program
  synthesis in genetic programming}. In \bibinfo{booktitle}{{\em GECCO '18:
  Proceedings of the Genetic and Evolutionary Computation Conference}}.
  \bibinfo{publisher}{ACM}, \bibinfo{address}{Kyoto, Japan},
  \bibinfo{pages}{1119--1126}.
\newblock
\showDOI{%
\url{https://doi.org/10.1145/3205455.3205592}}


\bibitem[\protect\citeauthoryear{Forstenlechner, Fagan, Nicolau, and
  O'Neill}{Forstenlechner et~al\mbox{.}}{2018c}]%
        {Forstenlechner:2018:CEC}
\bibfield{author}{\bibinfo{person}{Stefan Forstenlechner},
  \bibinfo{person}{David Fagan}, \bibinfo{person}{Miguel Nicolau}, {and}
  \bibinfo{person}{Michael O'Neill}.} \bibinfo{year}{2018}\natexlab{c}.
\newblock \showarticletitle{Towards Understanding and Refining the General
  Program Synthesis Benchmark Suite with Genetic Programming}. In
  \bibinfo{booktitle}{{\em 2018 IEEE Congress on Evolutionary Computation
  (CEC)}}, \bibfield{editor}{\bibinfo{person}{Marley Vellasco}} (Ed.).
  \bibinfo{publisher}{IEEE}, \bibinfo{address}{Rio de Janeiro, Brazil}.
\newblock


\bibitem[\protect\citeauthoryear{Helmuth, McPhee, Pantridge, and
  Spector}{Helmuth et~al\mbox{.}}{2017}]%
        {Helmuth:2017:GECCO}
\bibfield{author}{\bibinfo{person}{Thomas Helmuth},
  \bibinfo{person}{Nicholas~Freitag McPhee}, \bibinfo{person}{Edward
  Pantridge}, {and} \bibinfo{person}{Lee Spector}.}
  \bibinfo{year}{2017}\natexlab{}.
\newblock \showarticletitle{Improving Generalization of Evolved Programs
  Through Automatic Simplification}. In \bibinfo{booktitle}{{\em Proceedings of
  the Genetic and Evolutionary Computation Conference}} {\em
  (\bibinfo{series}{GECCO '17})}. \bibinfo{publisher}{ACM},
  \bibinfo{address}{Berlin, Germany}, \bibinfo{pages}{937--944}.
\newblock
\showDOI{%
\url{https://doi.org/10.1145/3071178.3071330}}


\bibitem[\protect\citeauthoryear{Helmuth, McPhee, and Spector}{Helmuth
  et~al\mbox{.}}{2015a}]%
        {Helmuth:2015:GPTP}
\bibfield{author}{\bibinfo{person}{Thomas Helmuth},
  \bibinfo{person}{Nicholas~Freitag McPhee}, {and} \bibinfo{person}{Lee
  Spector}.} \bibinfo{year}{2015}\natexlab{a}.
\newblock \showarticletitle{Lexicase Selection For Program Synthesis: A
  Diversity Analysis}. In \bibinfo{booktitle}{{\em Genetic Programming Theory
  and Practice XIII}} {\em (\bibinfo{series}{Genetic and Evolutionary
  Computation})}, \bibfield{editor}{\bibinfo{person}{Rick Riolo},
  \bibinfo{person}{William~P. Worzel}, \bibinfo{person}{M.~Kotanchek}, {and}
  \bibinfo{person}{A.~Kordon}} (Eds.). \bibinfo{publisher}{Springer},
  \bibinfo{address}{Ann Arbor, USA}.
\newblock
\showDOI{%
\url{https://doi.org/10.1007/978-3-319-34223-8}}


\bibitem[\protect\citeauthoryear{Helmuth, McPhee, and Spector}{Helmuth
  et~al\mbox{.}}{2016a}]%
        {Helmuth:2016:GECCOcomp}
\bibfield{author}{\bibinfo{person}{Thomas Helmuth},
  \bibinfo{person}{Nicholas~Freitag McPhee}, {and} \bibinfo{person}{Lee
  Spector}.} \bibinfo{year}{2016}\natexlab{a}.
\newblock \showarticletitle{Effects of Lexicase and Tournament Selection on
  Diversity Recovery and Maintenance}. In \bibinfo{booktitle}{{\em GECCO '16
  Companion: Proceedings of the Companion Publication of the 2016 Annual
  Conference on Genetic and Evolutionary Computation}}.
  \bibinfo{publisher}{ACM}, \bibinfo{address}{Denver, Colorado, USA},
  \bibinfo{pages}{983--990}.
\newblock
\showDOI{%
\url{https://doi.org/10.1145/2908961.2931657}}


\bibitem[\protect\citeauthoryear{Helmuth, McPhee, and Spector}{Helmuth
  et~al\mbox{.}}{2016b}]%
        {Helmuth:2016:GECCO}
\bibfield{author}{\bibinfo{person}{Thomas Helmuth},
  \bibinfo{person}{Nicholas~Freitag McPhee}, {and} \bibinfo{person}{Lee
  Spector}.} \bibinfo{year}{2016}\natexlab{b}.
\newblock \showarticletitle{The Impact of Hyperselection on Lexicase
  Selection}. In \bibinfo{booktitle}{{\em GECCO '16: Proceedings of the 2016
  Annual Conference on Genetic and Evolutionary Computation}},
  \bibfield{editor}{\bibinfo{person}{Tobias Friedrich}} (Ed.).
  \bibinfo{publisher}{ACM}, \bibinfo{address}{Denver, USA},
  \bibinfo{pages}{717--724}.
\newblock
\showDOI{%
\url{https://doi.org/10.1145/2908812.2908851}}


\bibitem[\protect\citeauthoryear{Helmuth and Spector}{Helmuth and
  Spector}{2013}]%
        {Helmuth:2013:GECCOcomp}
\bibfield{author}{\bibinfo{person}{Thomas Helmuth} {and} \bibinfo{person}{Lee
  Spector}.} \bibinfo{year}{2013}\natexlab{}.
\newblock \showarticletitle{Evolving a digital multiplier with the {PushGP}
  genetic programming system}. In \bibinfo{booktitle}{{\em GECCO '13 Companion:
  Proceeding of the fifteenth annual conference companion on Genetic and
  evolutionary computation conference companion}}. \bibinfo{publisher}{ACM},
  \bibinfo{address}{Amsterdam, The Netherlands}, \bibinfo{pages}{1627--1634}.
\newblock
\showDOI{%
\url{https://doi.org/10.1145/2464576.2466814}}


\bibitem[\protect\citeauthoryear{Helmuth and Spector}{Helmuth and
  Spector}{2015}]%
        {Helmuth:2015:GECCO}
\bibfield{author}{\bibinfo{person}{Thomas Helmuth} {and} \bibinfo{person}{Lee
  Spector}.} \bibinfo{year}{2015}\natexlab{}.
\newblock \showarticletitle{General Program Synthesis Benchmark Suite}. In
  \bibinfo{booktitle}{{\em GECCO '15: Proceedings of the 2015 Annual Conference
  on Genetic and Evolutionary Computation}}. \bibinfo{publisher}{ACM},
  \bibinfo{address}{Madrid, Spain}, \bibinfo{pages}{1039--1046}.
\newblock
\showDOI{%
\url{https://doi.org/10.1145/2739480.2754769}}


\bibitem[\protect\citeauthoryear{Helmuth, Spector, and Matheson}{Helmuth
  et~al\mbox{.}}{2015b}]%
        {Helmuth:2015:ieeeTEC}
\bibfield{author}{\bibinfo{person}{Thomas Helmuth}, \bibinfo{person}{Lee
  Spector}, {and} \bibinfo{person}{James Matheson}.}
  \bibinfo{year}{2015}\natexlab{b}.
\newblock \showarticletitle{Solving Uncompromising Problems with Lexicase
  Selection}.
\newblock \bibinfo{journal}{{\em IEEE Transactions on Evolutionary
  Computation\/}} \bibinfo{volume}{19}, \bibinfo{number}{5}
  (\bibinfo{date}{Oct.} \bibinfo{year}{2015}), \bibinfo{pages}{630--643}.
\newblock
\showISSN{1089-778X}
\showDOI{%
\url{https://doi.org/10.1109/TEVC.2014.2362729}}


\bibitem[\protect\citeauthoryear{Helmuth}{Helmuth}{2015}]%
        {Helmuth:thesis}
\bibfield{author}{\bibinfo{person}{Thomas~M. Helmuth}.}
  \bibinfo{year}{2015}\natexlab{}.
\newblock {\em \bibinfo{title}{General Program Synthesis from Examples Using
  Genetic Programming with Parent Selection Based on Random Lexicographic
  Orderings of Test Cases}}.
\newblock \bibinfo{thesistype}{Ph.D. Dissertation}. \bibinfo{school}{College of
  Information and Computer Sciences, University of Massachusetts Amherst},
  \bibinfo{address}{USA}.
\newblock
\showURL{%
\url{https://web.cs.umass.edu/publication/docs/2015/UM-CS-PhD-2015-005.pdf}}


\bibitem[\protect\citeauthoryear{Jackson}{Jackson}{2010}]%
        {Jackson:2010:PPSN}
\bibfield{author}{\bibinfo{person}{David Jackson}.}
  \bibinfo{year}{2010}\natexlab{}.
\newblock \showarticletitle{Promoting Phenotypic Diversity in Genetic
  Programming}. In \bibinfo{booktitle}{{\em PPSN 2010 11th International
  Conference on Parallel Problem Solving From Nature}} {\em
  (\bibinfo{series}{Lecture Notes in Computer Science})},
  \bibfield{editor}{\bibinfo{person}{Robert Schaefer}, \bibinfo{person}{Carlos
  Cotta}, \bibinfo{person}{Joanna Kolodziej}, {and} \bibinfo{person}{Guenter
  Rudolph}} (Eds.), Vol.~\bibinfo{volume}{6239}. \bibinfo{publisher}{Springer},
  \bibinfo{address}{Krakow, Poland}, \bibinfo{pages}{472--481}.
\newblock


\bibitem[\protect\citeauthoryear{{La Cava}, Helmuth, Spector, and Moore}{{La
  Cava} et~al\mbox{.}}{2018}]%
        {LaCava:EC}
\bibfield{author}{\bibinfo{person}{William {La Cava}}, \bibinfo{person}{Thomas
  Helmuth}, \bibinfo{person}{Lee Spector}, {and} \bibinfo{person}{Jason~H.
  Moore}.} \bibinfo{year}{2018}\natexlab{}.
\newblock \showarticletitle{A probabilistic and multi-objective analysis of
  lexicase selection and epsilon-lexicase selection}.
\newblock \bibinfo{journal}{{\em Evolutionary Computation\/}}
  (\bibinfo{year}{2018}).
\newblock
\showISSN{1063-6560}
\newblock
\shownote{Forthcoming.}


\bibitem[\protect\citeauthoryear{{La Cava}, Spector, and Danai}{{La Cava}
  et~al\mbox{.}}{2016}]%
        {LaCava:2016:GECCO}
\bibfield{author}{\bibinfo{person}{William {La Cava}}, \bibinfo{person}{Lee
  Spector}, {and} \bibinfo{person}{Kourosh Danai}.}
  \bibinfo{year}{2016}\natexlab{}.
\newblock \showarticletitle{Epsilon-lexicase Selection for Regression}. In
  \bibinfo{booktitle}{{\em GECCO '16: Proceedings of the 2016 Annual Conference
  on Genetic and Evolutionary Computation}},
  \bibfield{editor}{\bibinfo{person}{Tobias Friedrich}} (Ed.).
  \bibinfo{publisher}{ACM}, \bibinfo{address}{Denver, USA},
  \bibinfo{pages}{741--748}.
\newblock
\showDOI{%
\url{https://doi.org/10.1145/2908812.2908898}}


\bibitem[\protect\citeauthoryear{Liskowski, Krawiec, Helmuth, and
  Spector}{Liskowski et~al\mbox{.}}{2015}]%
        {Liskowski:2015:GECCOcomp}
\bibfield{author}{\bibinfo{person}{Pawel Liskowski}, \bibinfo{person}{Krzysztof
  Krawiec}, \bibinfo{person}{Thomas Helmuth}, {and} \bibinfo{person}{Lee
  Spector}.} \bibinfo{year}{2015}\natexlab{}.
\newblock \showarticletitle{Comparison of Semantic-aware Selection Methods in
  Genetic Programming}. In \bibinfo{booktitle}{{\em GECCO 2015 Semantic Methods
  in Genetic Programming (SMGP'15) Workshop}}. \bibinfo{publisher}{ACM},
  \bibinfo{address}{Madrid, Spain}, \bibinfo{pages}{1301--1307}.
\newblock
\showDOI{%
\url{https://doi.org/10.1145/2739482.2768505}}


\bibitem[\protect\citeauthoryear{McPhee, Donatucci, and Helmuth}{McPhee
  et~al\mbox{.}}{2015}]%
        {McPhee:2015:GPTP}
\bibfield{author}{\bibinfo{person}{Nicholas~Freitag McPhee},
  \bibinfo{person}{David Donatucci}, {and} \bibinfo{person}{Thomas Helmuth}.}
  \bibinfo{year}{2015}\natexlab{}.
\newblock \showarticletitle{Using Graph Databases to Explore Genetic
  Programming Run Dynamics}. In \bibinfo{booktitle}{{\em Genetic Programming
  Theory and Practice XIII}} {\em (\bibinfo{series}{Genetic and Evolutionary
  Computation})}. \bibinfo{publisher}{Springer}, \bibinfo{address}{Ann Arbor,
  USA}.
\newblock
\showURL{%
\url{http://www.springer.com/us/book/9783319342214}}


\bibitem[\protect\citeauthoryear{McPhee, Helmuth, and Spector}{McPhee
  et~al\mbox{.}}{2017}]%
        {McPhee:2017:GECCOa}
\bibfield{author}{\bibinfo{person}{Nicholas~Freitag McPhee},
  \bibinfo{person}{Thomas Helmuth}, {and} \bibinfo{person}{Lee Spector}.}
  \bibinfo{year}{2017}\natexlab{}.
\newblock \showarticletitle{Using Algorithm Configuration Tools to Optimize
  Genetic Programming Parameters: A Case Study}. In \bibinfo{booktitle}{{\em
  Proceedings of the Genetic and Evolutionary Computation Conference
  Companion}} {\em (\bibinfo{series}{GECCO '17})}. \bibinfo{publisher}{ACM},
  \bibinfo{address}{Berlin, Germany}, \bibinfo{pages}{243--244}.
\newblock
\showDOI{%
\url{https://doi.org/10.1145/3067695.3076097}}


\bibitem[\protect\citeauthoryear{Metevier, Saini, and Spector}{Metevier
  et~al\mbox{.}}{2019}]%
        {Metevier2019}
\bibfield{author}{\bibinfo{person}{Blossom Metevier},
  \bibinfo{person}{Anil~Kumar Saini}, {and} \bibinfo{person}{Lee Spector}.}
  \bibinfo{year}{2019}\natexlab{}.
\newblock \bibinfo{booktitle}{{\em Lexicase Selection Beyond Genetic
  Programming}}.
\newblock \bibinfo{publisher}{Springer International Publishing},
  \bibinfo{address}{Cham}, \bibinfo{pages}{123--136}.
\newblock
\showISBNx{978-3-030-04735-1}
\showDOI{%
\url{https://doi.org/10.1007/978-3-030-04735-1_7}}


\bibitem[\protect\citeauthoryear{Moore and Stanton}{Moore and Stanton}{2017}]%
        {moore:2017:ecal}
\bibfield{author}{\bibinfo{person}{Jared~M. Moore} {and} \bibinfo{person}{Adam
  Stanton}.} \bibinfo{year}{2017}\natexlab{}.
\newblock \showarticletitle{Lexicase selection outperforms previous strategies
  for incremental evolution of virtual creature controllers}.
\newblock \bibinfo{journal}{{\em Proceedings of the European Conference on
  Artificial Life\/}} (\bibinfo{year}{2017}), \bibinfo{pages}{290--297}.
\newblock
\showDOI{%
\url{https://doi.org/10.1162/ecal\_a\_0050\_14}}


\bibitem[\protect\citeauthoryear{Pantridge, Helmuth, McPhee, and
  Spector}{Pantridge et~al\mbox{.}}{2018}]%
        {Pantridge:2018:GECCO:SEL}
\bibfield{author}{\bibinfo{person}{Edward Pantridge}, \bibinfo{person}{Thomas
  Helmuth}, \bibinfo{person}{Nicholas~Freitag McPhee}, {and}
  \bibinfo{person}{Lee Spector}.} \bibinfo{year}{2018}\natexlab{}.
\newblock \showarticletitle{Specialization and Elitism in Lexicase and
  Tournament Selection}. In \bibinfo{booktitle}{{\em Proceedings of the Genetic
  and Evolutionary Computation Conference Companion}} {\em
  (\bibinfo{series}{GECCO '18})}. \bibinfo{publisher}{ACM},
  \bibinfo{address}{New York, NY, USA}, \bibinfo{pages}{1914--1917}.
\newblock
\showISBNx{978-1-4503-5764-7}
\showDOI{%
\url{https://doi.org/10.1145/3205651.3208220}}


\bibitem[\protect\citeauthoryear{Pantridge and Spector}{Pantridge and
  Spector}{2017}]%
        {Pantridge:2017:GECCO}
\bibfield{author}{\bibinfo{person}{Edward Pantridge} {and} \bibinfo{person}{Lee
  Spector}.} \bibinfo{year}{2017}\natexlab{}.
\newblock \showarticletitle{{PyshGP}: {PushGP} in Python}. In
  \bibinfo{booktitle}{{\em Proceedings of the Genetic and Evolutionary
  Computation Conference Companion}} {\em (\bibinfo{series}{GECCO '17})}.
  \bibinfo{publisher}{ACM}, \bibinfo{address}{Berlin, Germany},
  \bibinfo{pages}{1255--1262}.
\newblock
\showDOI{%
\url{https://doi.org/10.1145/3067695.3082468}}


\bibitem[\protect\citeauthoryear{Rosin}{Rosin}{2018}]%
        {DBLP:journals/corr/abs-1811-10665}
\bibfield{author}{\bibinfo{person}{Christopher~D. Rosin}.}
  \bibinfo{year}{2018}\natexlab{}.
\newblock \showarticletitle{Stepping Stones to Inductive Synthesis of Low-Level
  Looping Programs}.
\newblock \bibinfo{journal}{{\em CoRR\/}}  \bibinfo{volume}{abs/1811.10665}
  (\bibinfo{year}{2018}).
\newblock
\showeprint[arxiv]{1811.10665}
\showURL{%
\url{http://arxiv.org/abs/1811.10665}}


\bibitem[\protect\citeauthoryear{Spector}{Spector}{2012}]%
        {Spector:2012:GECCOcompA}
\bibfield{author}{\bibinfo{person}{Lee Spector}.}
  \bibinfo{year}{2012}\natexlab{}.
\newblock \showarticletitle{Assessment of Problem Modality by Differential
  Performance of Lexicase Selection in Genetic Programming: A Preliminary
  Report}. In \bibinfo{booktitle}{{\em 1st workshop on Understanding Problems
  (GECCO-UP)}}, \bibfield{editor}{\bibinfo{person}{Kent McClymont} {and}
  \bibinfo{person}{Ed~Keedwell}} (Eds.). \bibinfo{publisher}{ACM},
  \bibinfo{address}{Philadelphia, Pennsylvania, USA},
  \bibinfo{pages}{401--408}.
\newblock
\showDOI{%
\url{https://doi.org/10.1145/2330784.2330846}}


\bibitem[\protect\citeauthoryear{Spector, Klein, and Keijzer}{Spector
  et~al\mbox{.}}{2005}]%
        {1068292}
\bibfield{author}{\bibinfo{person}{Lee Spector}, \bibinfo{person}{Jon Klein},
  {and} \bibinfo{person}{Maarten Keijzer}.} \bibinfo{year}{2005}\natexlab{}.
\newblock \showarticletitle{The Push3 execution stack and the evolution of
  control}. In \bibinfo{booktitle}{{\em {GECCO 2005}: Proceedings of the 2005
  conference on Genetic and evolutionary computation}},
  Vol.~\bibinfo{volume}{2}. \bibinfo{publisher}{ACM Press},
  \bibinfo{address}{Washington DC, USA}, \bibinfo{pages}{1689--1696}.
\newblock
\showISBNx{1-59593-010-8}
\showDOI{%
\url{https://doi.org/10.1145/1068009.1068292}}


\bibitem[\protect\citeauthoryear{Spector, La~Cava, Shanabrook, Helmuth, and
  Pantridge}{Spector et~al\mbox{.}}{2018}]%
        {10.1007/978-3-319-90512-9_7}
\bibfield{author}{\bibinfo{person}{Lee Spector}, \bibinfo{person}{William
  La~Cava}, \bibinfo{person}{Saul Shanabrook}, \bibinfo{person}{Thomas
  Helmuth}, {and} \bibinfo{person}{Edward Pantridge}.}
  \bibinfo{year}{2018}\natexlab{}.
\newblock \showarticletitle{Relaxations of Lexicase Parent Selection}. In
  \bibinfo{booktitle}{{\em Genetic Programming Theory and Practice XV}},
  \bibfield{editor}{\bibinfo{person}{Wolfgang Banzhaf},
  \bibinfo{person}{Randal~S. Olson}, \bibinfo{person}{William Tozier}, {and}
  \bibinfo{person}{Rick Riolo}} (Eds.). \bibinfo{publisher}{Springer
  International Publishing}, \bibinfo{address}{Cham},
  \bibinfo{pages}{105--120}.
\newblock
\showISBNx{978-3-319-90512-9}


\bibitem[\protect\citeauthoryear{Spector and Robinson}{Spector and
  Robinson}{2002}]%
        {spector:2002:GPEM}
\bibfield{author}{\bibinfo{person}{Lee Spector} {and} \bibinfo{person}{Alan
  Robinson}.} \bibinfo{year}{2002}\natexlab{}.
\newblock \showarticletitle{Genetic Programming and Autoconstructive Evolution
  with the Push Programming Language}.
\newblock \bibinfo{journal}{{\em Genetic Programming and Evolvable Machines\/}}
  \bibinfo{volume}{3}, \bibinfo{number}{1} (\bibinfo{date}{March}
  \bibinfo{year}{2002}), \bibinfo{pages}{7--40}.
\newblock
\showISSN{1389-2576}
\showDOI{%
\url{https://doi.org/10.1023/A:1014538503543}}


\end{thebibliography}

\newpage
\thispagestyle{empty}
\onecolumn

\section*{Erratum Notice}

After publication, it came to our attention that there were errors in the data presented in Figure~\ref{figure:es:case-usage-distrib}. Those errors are corrected in Figure~\ref{figure:es:case-usage-distrib} in this PDF. These corrections do not influence the discussion presented in the text, and therefore the text has not been changed. The originally published and incorrect version of the figure can be found below.

\begin{figure}[H]
    \centering
    \includegraphics[width=0.9\linewidth]{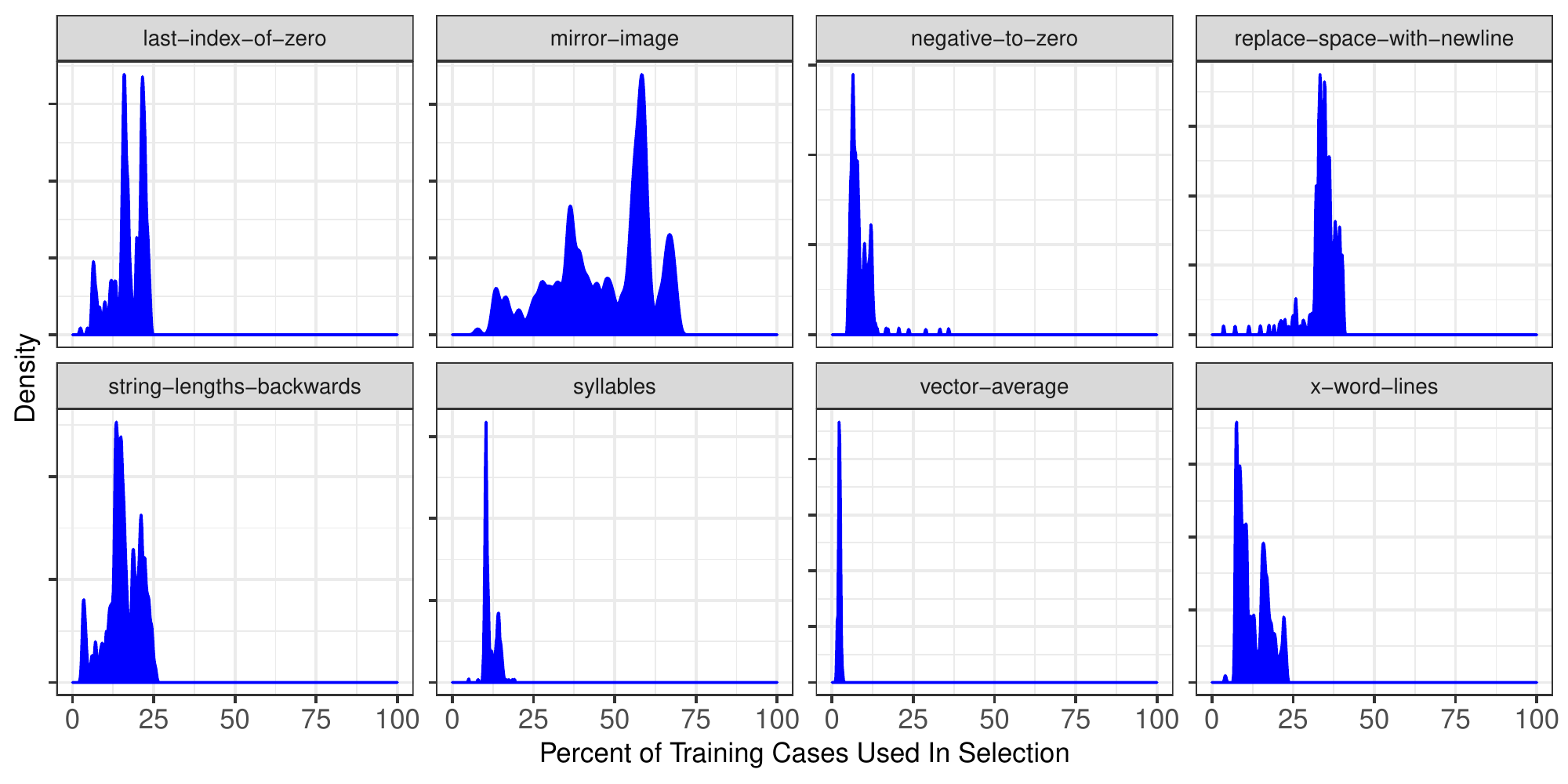}
\end{figure}

\end{document}